\documentclass[journal, 12pt, draftclsnofoot, onecolumn]{IEEEtran}
\usepackage{pgfplots}

\usepackage{cite}

  \usepackage{graphicx}
  \graphicspath{{figs/}}
  \DeclareGraphicsExtensions{.eps}
\usepackage{amsmath}
\ifCLASSOPTIONcompsoc
  \usepackage[caption=false,font=normalsize,labelfont=sf,textfont=sf]{subfig}
\else
  \usepackage[caption=false,font=footnotesize]{subfig}
\fi
\usepackage{fixltx2e}
\usepackage{url}

\begin{document}
%
\title{Online Keyword Spotting with a\\Character-Level Recurrent Neural Network}
%
%
%

\author{Kyuyeon~Hwang,~\IEEEmembership{Student~Member,~IEEE,}
        Minjae~Lee,~\IEEEmembership{Student~Member,~IEEE,}
        and~Wonyong~Sung,~\IEEEmembership{Fellow,~IEEE}
\thanks{K. Hwang, M. Lee, and W. Sung are with the Department
of Electrical and Computer Engineering, Seoul National University, Seoul,
08826 Korea (e-mail: \texttt{kyuyeon.hwang@gmail.com; mjlee@dsp.snu.ac.kr; wysung@snu.ac.kr}).}
}
\maketitle

\begin{abstract}
In this paper, we propose a context-aware keyword spotting model employing a character-level recurrent neural network (RNN) for spoken term detection in continuous speech. The RNN is end-to-end trained with connectionist temporal classification (CTC) to generate the probabilities of character and word-boundary labels. There is no need for the phonetic transcription, senone modeling, or system dictionary in training and testing. Also, keywords can easily be added and modified by editing the text based keyword list without retraining the RNN. Moreover, the unidirectional RNN processes an infinitely long input audio streams without pre-segmentation and keywords are detected with low-latency before the utterance is finished. Experimental results show that the proposed keyword spotter significantly outperforms the deep neural network (DNN) and hidden Markov model (HMM) based keyword-filler model even with less computations.
\end{abstract}


%
\IEEEpeerreviewmaketitle

\section{Introduction}
\label{sec:intro}

\IEEEPARstart{K}{eyword} spotting, or spoken term detection, is a technique for detecting pre-defined keywords in continuous speech. This technique is used for numerous applications including voice activation of smartphones, voice command for TVs and automobiles, and audio search. Compared to large vocabulary continuous speech recognition (LVCSR), keyword spotting is usually designed with a simple structure, which is suitable for real-time low-power systems.

The traditional hidden Markov model (HMM) based keyword spotter \cite{rose1990hidden} does not include a language model to limit the complexity of the decoding stage. Therefore, the algorithm only makes use of the pronunciation without knowing the context of the utterances. This is especially problematic when the phonetic description of a certain keyword is included in other longer spoken words, as there is no way to distinguish whether the detected phonetic sequence is actually belonging to the matching keyword or a part of other words. For example, the keyword ``honey" can be detected when ``honeymoon" is spoken.

To remedy this issue, a recurrent neural network (RNN) based context-aware keyword spotter is proposed in this paper. The RNN is end-to-end trained with connectionist temporal classification (CTC) \cite{graves2006connectionist} to directly transcribe the input speech to a sequence of character labels and a word-boundary label. The word-boundary label plays a key role in filtering out the case where a certain keyword string is included in other words. The soft output of the RNN is fed into a simple decoding network for computing posterior probabilities of keywords.

The proposed keyword spotter has several advantages over the previous approaches as follows:
\begin{itemize}
\item{
The front-end RNN is unidirectional and trained by online CTC \cite{hwang2015online} to process infinitely long audio streams with low latency. Therefore, there is no need for pre-segmenting the utterance with an additional voice activity detector \cite{sohn1999statistical}. On the other hand, in the previous approaches with bidirectional RNNs \cite{fernandez2007application, wollmer2010bidirectional}, keywords are detected after listening each (pre-segmented) utterance to the end.}
\item{
The RNN is end-to-end trained. Unlike the previous phoneme based approaches \cite{rose1990hidden, wollmer2010bidirectional}, there is no need for phoneme or senone modeling in both learning and testing. Also, the RNN learns vocabulary and weak language models, which allows it to detect word boundaries and greatly improve the accuracy in continuous speech.}
\item{
New keywords can be added without retraining the RNN by modifying the back-end that is simpler than that of the HMM keyword-filler model \cite{rose1990hidden}. This is not possible with the previous end-to-end approaches \cite{fernandez2007application, chen2014small}.}
\end{itemize}

Experimental results show that the proposed CTC keyword spotter greatly outperforms the deep neural network (DNN) \cite{hinton2012deep} and HMM \cite{rose1990hidden} hybrid keyword-filler model on the Wall Street Journal (WSJ) \cite{paul1992design} Nov'92 20K evaluation set especially when the length of the keyword is short.

\section{End-to-End Keyword Spotting Model}

The proposed keyword spotter consists of a front-end RNN and a back-end decoder. At each frame, the RNN converts an input audio feature to probabilities of characters. Then, the decoder computes probabilities of the keywords based on the recent character-level probabilities.

\subsection{Character-Level Unidirectional RNN}
\label{ssec:front-end}

The front-end RNN is a deep unidirectional long short-term memory (LSTM) \cite{hochreiter1997long} network with forget gates \cite{gers2000learning} and peephole connections \cite{gers2003learning}. Specifically, the RNN consists of three LSTM layers and a softmax output layer \cite{bridle1990probabilistic}, and sequence-to-sequence trained with connectionist temporal classification (CTC) \cite{graves2006connectionist}. The network is similar to the one employed for the end-to-end speech recognition \cite{graves2014towards}. However, in our case, the network is unidirectional and trained with online CTC \cite{hwang2015online} on very long speech streams, instead of bidirectional LSTM networks with sequence-wise standard CTC training. This enables the RNN to process infinitely long input speech.

CTC \cite{graves2006connectionist} is one of the most successful sequence-to-sequence learning algorithms for RNNs. Recently, CTC based end-to-end trained RNNs have been reported to show impressive performance in automatic speech recognition \cite{graves2014towards, amodei2015deep}, which are comparable to state-of-the-art DNN-HMM models. In those end-to-end models, RNNs learn how to directly transcribe the input speech to the corresponding character-level transcription without intermediate senone or phoneme modeling. Moreover, the RNNs can attain weak language models from the training speech, which diminishes the need for external lexicon models or word dictionaries. Although the language modeling capability is not sufficient for the application to LVCSR, this improves the accuracy of keyword spotting very much.

For the proposed keyword spotter, we employ the online CTC algorithm \cite{hwang2015online} to train continuously running unidirectional RNNs, which can process an infinitely long input speech without pre-segmentation. For this, the training is performed on the infinitely long training streams that are generated by randomly concatenating training sequences. The online CTC algorithm allows the RNN to learn sequences that are longer than the unroll amount. Therefore, a fixed amount of unroll can be used for training sequences with various lengths. This enables efficient RNN training with multiple parallel training streams on a GPU with synchronized forward and backward steps \cite{hwang2015single} using the online backpropagation through time algorithm \cite{williams1990efficient}. The networks are unrolled 2,048 times and weight updates are performed every 1,024 forward steps with stochastic gradient descent (SGD).

As the input to the RNN, 40-dimensional log-filterbank features plus energy and their delta and double-delta (total 123-dimensional vector) are used. The feature vectors are extracted using a 25~ms Hamming window with 10~ms period and element-wisely normalized to zero mean and unit standard deviation based on the statistics obtained from the training set. The output is the probabilities for 30 labels including 26 alphabets, 2 special characters ('~and~.), the word-boundary label (\_), and the CTC blank label (-). The CTC blank label is generated when there is no specific output to emit.

All speaker independent training utterances in WSJ corpus \cite{paul1992design} except verbalized punctuation versions are used for training. Also, odd transcriptions are filtered out, which makes the final training set contain 167 hours of speech. The WSJ Nov'93 20K development set is used as the validation set for annealing and early stopping.

\subsection{Decoder Models}
\label{ssec:back-end}

\begin{figure}[t]
    \begin{center}
    \subfloat[]{
    \includegraphics[width=3.0 in]{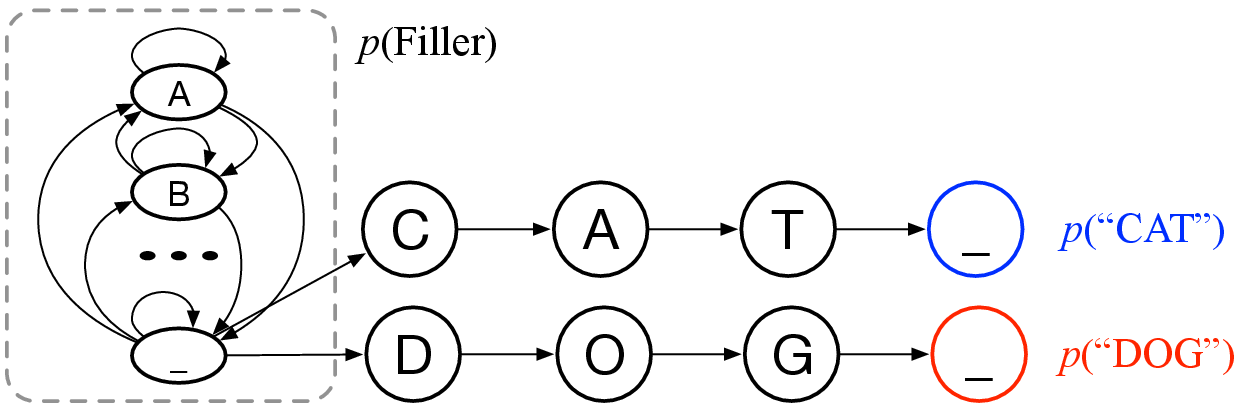}%
    \label{fig:keyword-filler}
    }
    \end{center}
    \begin{center}
    \subfloat[]{
    \includegraphics[width=3.0 in]{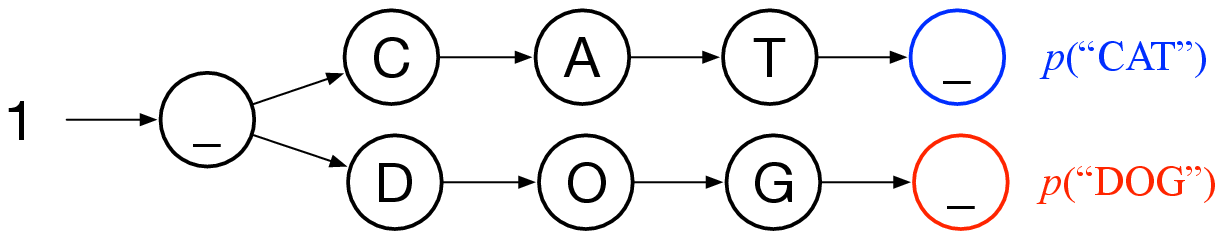}%
    \label{fig:keyword}
    }
    \end{center}

\caption{(a) CTC keyword-filler and (b) keyword-only decoding network.}

\end{figure}

\begin{figure}[t]
\begin{center}
\includegraphics[width=2.5 in]{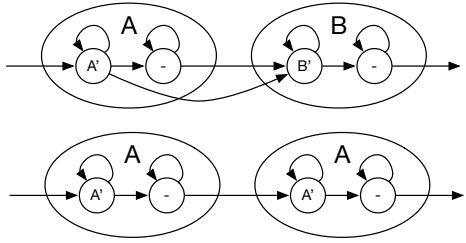}%
\end{center}
\caption{CTC state transition diagram where ``-" indicates the CTC blank label.}
\label{fig:ctc_state}
\end{figure}


The output of the CTC-trained network is a vector that contains the probabilities of labels (i.e., characters, the word-boundary label, and the CTC blank label). The goal of decoding is to convert the character-level output sequence to the sequence of keyword-level probabilities.

To spot keywords with the soft output of the RNN, two back-end decoders are considered in this paper. As shown in \figurename~\ref{fig:keyword-filler}, the first one is similar to the HMM keyword spotter with a filler model \cite{rose1990hidden}. Instead of HMM states, character-level nodes are used for filler and keyword networks. The keyword network starts and ends with the word-boundary labels, ``\_". At every frame, the posterior probability of the filler model and the keyword model are compared. When the difference of the two log-posterior probabilities is below a threshold, the keyword is detected. For discriminative keyword spotting, multiple keyword networks can be employed. The definition of the keyword and filler probabilities will be described later in this Section.

The second decoding network is depicted in \figurename~\ref{fig:keyword}, which is similar to the first one, but the filler network is removed. This is equivalent to setting the posterior probability of the filler model to one and inserting a word-boundary node before the keyword network. A keyword is detected when the negative log-posterior probability is below a threshold. The experiments in Section~\ref{sec:experiments} show that there is virtually no performance difference between the two decoding networks. That is, the filler network is not needed for the CTC keyword spotter.

Actual state transition between two label nodes exactly follows the original formulation of the standard CTC. As depicted in \figurename~\ref{fig:ctc_state}, each node in \figurename~\ref{fig:keyword-filler} and \figurename~\ref{fig:keyword} consists of a character label or the word-boundary label followed by the CTC blank label. Note that transitions between the same CTC labels are not allowed. At every frame, each state updates its value by multiplying the net incoming probability by the corresponding RNN output, where the net probability is the sum of all incoming probabilities. However, when the filler model is employed, the incoming probabilities may have different label histories. In this case, the exact net probability cannot be obtained since the probabilities from different paths cannot be summed. Therefore, we approximate the sum operation to the max operation to find the best alignment (with timing) instead of the best path (without timing). Experimental results show that this approximation does not degrade the decoding accuracy when applied to the keyword-only model.

The probability of the filler model is defined as the maximum posterior probability among all CTC states inside the filler model. On the other hand, the probability of the keyword network is the summed (maximum, if the sum-to-max approximation is applied) posterior probability of the two CTC states inside the last word-boundary node (red and blue circles in \figurename~\ref{fig:keyword-filler} and \figurename~\ref{fig:keyword}). Precision-recall trade-off is adjusted by the threshold value between these two probabilities. Setting a large threshold increases the recall at the cost of the reduced precision, and vice versa.

\section{Experiments}
\label{sec:experiments}

\subsection{Baseline DNN-HMM Keyword-Filler Model}

As a baseline, the performance of the DNN-HMM keyword spotter is also obtained, where the Gaussian mixture model (GMM) of the keyword-filler model in \cite{rose1990hidden} is replaced by a DNN acoustic model \cite{hinton2012deep}. The DNN has 4 layers with 1,024 logistic sigmoid units each and a softmax layer with 2,717 units, which corresponds to 2,717 tied triphone states. The input of the DNN-HMM keyword spotter is 17 consecutive 39-dimensional Mel-frequency cepstral coefficient (MFCC) feature vectors (13 cepstral coefficients plus their delta and double-delta values) extracted at every 10~ms with 25~ms Hamming window. It is observed in our DNN-HMM experiments that the MFCC feature performs better than the log-filterbank features. The baseline model is trained on the same 167 hours of training set that is used for the proposed model.

\subsection{Experimental Setup}

The experiments are performed on the 42-minute audio stream that is generated by concatenating all 330 utterances in the WSJ Nov'92 20K evaluation set.

For the DNN-HMM keyword spotter with multiple keywords, the threshold of each keyword is given to be proportional to the duration of the keyword pronounced. For the proposed CTC keyword spotter, on the other hand, the threshold is proportional to the number of characters in the keyword.

The models are evaluated on the following keyword sets:
\begin{itemize}
\item{
\emph{Set A}: percent, hundred, thousand, million, people, president, average, foreign, international, nuclear}
\item{
\emph{Set B}: that, this, they, but, and, or, with, from, will, not}
\end{itemize}
Set A consists of multisyllabic words. On the other hand, Set B is a set of monosyllabic words, which are extremely difficult to detect without understanding the underlying linguistic structures.

\subsection{Evaluation}

\subsubsection{Detection Latency}

\begin{figure*}[t]
\begin{center}
\includegraphics[width=\textwidth]{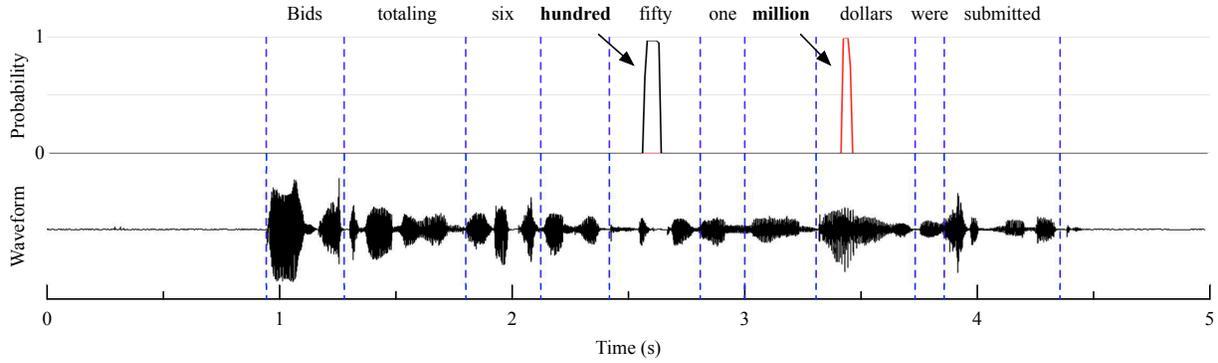}%
\end{center}
\caption{Input waveform and the output (posterior probabilities) of the keyword-only decoder with manually aligned word boundaries. The keywords are ``hundred" and ``million''. In this example, the detection latencies are observed to be less than 200~ms.}
\label{fig:timing}
\end{figure*}

The proposed keyword spotter employs unidirectional RNNs for low-latency online spoken term detection. When unidirectional networks are trained with CTC, the network learns the output delay that is required for making use of sufficient amount of information from the input. \figurename~\ref{fig:timing} shows the input waveform and the posterior probabilities of the keywords from the keyword-only decoder. In this example, the keywords are detected within 200~ms after they are completely spoken. This is comparable to the human reaction time to speech stimuli, which is also roughly 200~ms as reported in \cite{fry1975simple}.

\subsubsection{Comparison of the Decoding Models}

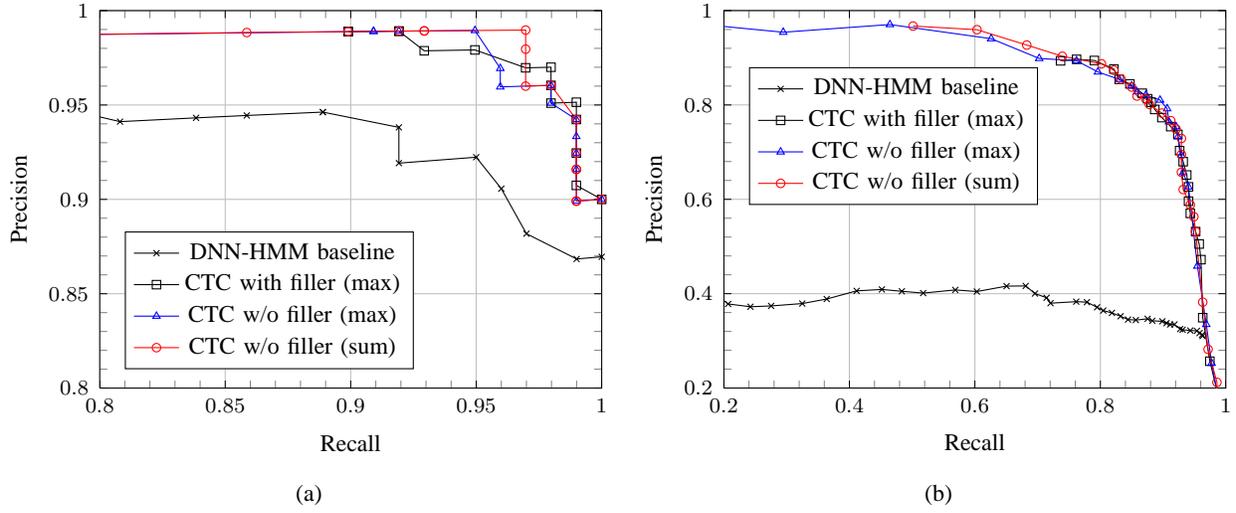
\begin{figure*}[t!]
    \centering
    \subfloat[]{%
{%
\begin{tikzpicture}
\begin{axis}
[
width= 0.5\textwidth,
height=0.4\textwidth,
compat=1.3,
xmin=0.8,
ymin=0.8,
xmax=1,
ymax=1,
label style={font=\footnotesize},
xlabel={Recall},
ylabel={Precision},
legend style={font=\footnotesize,at={(0.05,0.05)},anchor=south west},
tick label style={font=\scriptsize},
domain=1:512,
xmajorgrids,
ymajorgrids,
minor x tick num=4,
minor y tick num=4,
log basis x={10},
xtick pos=both,
xtick align=inside,
major tick style={line width=0.010cm, black},
major tick length=0.10cm
]%
\legend{DNN-HMM baseline, CTC with filler (max), CTC w/o filler (max), CTC w/o filler (sum)};
\addplot[color=black, mark=x, mark size=1.6pt, solid, mark repeat=1,mark options=solid]
file{data/set_a_dnn_hmm.txt};
\addplot[color=black, mark=square, mark size=1.6pt, solid, mark repeat=1,mark options=solid]
file{data/set_a_3x768_filler_max.txt};
\addplot[color=blue, mark=triangle, mark size=1.6pt, solid, mark repeat=1,mark options=solid]
file{data/set_a_3x768_no_filler_max.txt};
\addplot[color=red, mark=o, mark size=1.6pt, solid, mark repeat=1,mark options=solid]
file{data/set_a_3x768_no_filler_sum.txt};
\end{axis}%
\end{tikzpicture}%
}%

    }%
    ~ 
    \subfloat[]{%
{%
\begin{tikzpicture}
\begin{axis}
[
width= 0.5\textwidth,
height=0.4\textwidth,
compat=1.3,
xmin=0.2,
ymin=0.2,
xmax=1,
ymax=1,
label style={font=\footnotesize},
xlabel={Recall},
ylabel={Precision},
legend style={font=\footnotesize,at={(0.05,0.85)},anchor=north west},
tick label style={font=\scriptsize},
domain=1:512,
xmajorgrids,
ymajorgrids,
minor x tick num=4,
minor y tick num=4,
log basis x={10},
xtick pos=both,
xtick align=inside,
major tick style={line width=0.010cm, black},
major tick length=0.10cm
]%
\legend{DNN-HMM baseline, CTC with filler (max), CTC w/o filler (max), CTC w/o filler (sum)};
\addplot[color=black, mark=x, mark size=1.6pt, solid, mark repeat=1,mark options=solid]
file{data/set_b_dnn_hmm.txt};
\addplot[color=black, mark=square, mark size=1.6pt, solid, mark repeat=1,mark options=solid]
file{data/set_b_3x768_filler_max.txt};
\addplot[color=blue, mark=triangle, mark size=1.6pt, solid, mark repeat=1,mark options=solid]
file{data/set_b_3x768_no_filler_max.txt};
\addplot[color=red, mark=o, mark size=1.6pt, solid, mark repeat=1,mark options=solid]
file{data/set_b_3x768_no_filler_sum.txt};
\end{axis}%
\end{tikzpicture}%
}%
    }%
    \caption{Precision-recall plot for (a) Set A (multisyllabic keywords) and (b) Set B (monosyllabic keywords) with various decoding models.}
    \label{fig:plot_decoders}
\end{figure*}

The precision-recall plots for Set A and Set B are shown in \figurename~\ref{fig:plot_decoders}, with various decoding models where the network size is fixed to 3x768 (i.e., 3 LSTM layers with 768 cells each). The experiments are performed with the baseline DNN-HMM model, the CTC keyword-filler model, and the CTC keyword-only model. Also, we observe the effect of approximating the sum operations to the max operations for the CTC keyword-only model. As shown in the figures, the proposed CTC based keyword spotters outperform the DNN-HMM based one especially on the Set B. This indicates that the subsidiary information such as word boundaries, the linguistic structure, and context is very helpful for detecting the monosyllabic keywords. Note that there is no advantage of employing the filler model for the CTC keyword spotter. Moreover, the sum-to-max approximation does not degrade the detection accuracy. 

\subsubsection{Comparison of Various Network Sizes}

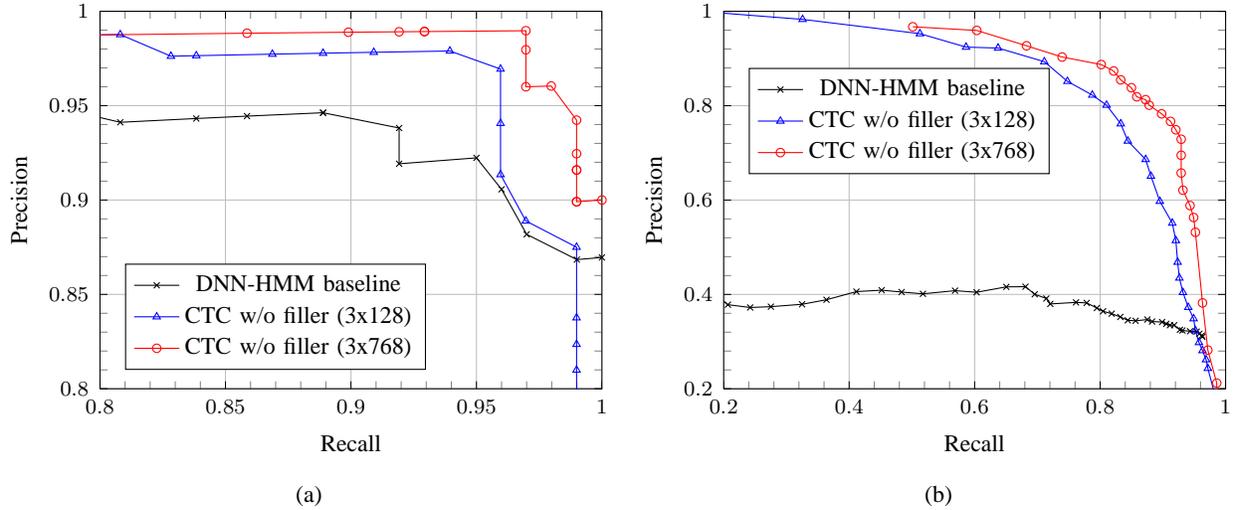
\begin{figure*}[t!]
    \centering
    \subfloat[]{%
{%
\begin{tikzpicture}
\begin{axis}
[
width= 0.5\textwidth,
height=0.4\textwidth,
compat=1.3,
xmin=0.8,
ymin=0.8,
xmax=1,
ymax=1,
label style={font=\footnotesize},
xlabel={Recall},
ylabel={Precision},
legend style={font=\footnotesize,at={(0.05,0.05)},anchor=south west},
tick label style={font=\scriptsize},
domain=1:512,
xmajorgrids,
ymajorgrids,
minor x tick num=4,
minor y tick num=4,
log basis x={10},
xtick pos=both,
xtick align=inside,
major tick style={line width=0.010cm, black},
major tick length=0.10cm
]%
\legend{DNN-HMM baseline, CTC w/o filler (3x128), CTC w/o filler (3x768), CTC w/o filler (sum)};
\addplot[color=black, mark=x, mark size=1.6pt, solid, mark repeat=1,mark options=solid]
file{data/set_a_dnn_hmm.txt};
\addplot[color=blue, mark=triangle, mark size=1.6pt, solid, mark repeat=1,mark options=solid]
file{data/set_a_3x128_no_filler_sum.txt};
\addplot[color=red, mark=o, mark size=1.6pt, solid, mark repeat=1,mark options=solid]
file{data/set_a_3x768_no_filler_sum.txt};
\end{axis}%
\end{tikzpicture}%
}%

    }%
    ~ 
    \subfloat[]{%
{%
\begin{tikzpicture}
\begin{axis}
[
width= 0.5\textwidth,
height=0.4\textwidth,
compat=1.3,
xmin=0.2,
ymin=0.2,
xmax=1,
ymax=1,
label style={font=\footnotesize},
xlabel={Recall},
ylabel={Precision},
legend style={font=\footnotesize,at={(0.05,0.85)},anchor=north west},
tick label style={font=\scriptsize},
domain=1:512,
xmajorgrids,
ymajorgrids,
minor x tick num=4,
minor y tick num=4,
log basis x={10},
xtick pos=both,
xtick align=inside,
major tick style={line width=0.010cm, black},
major tick length=0.10cm
]%
%
\legend{DNN-HMM baseline, CTC w/o filler (3x128), CTC w/o filler (3x768), CTC w/o filler (sum)};
\addplot[color=black, mark=x, mark size=1.6pt, solid, mark repeat=1,mark options=solid]
file{data/set_b_dnn_hmm.txt};
\addplot[color=blue, mark=triangle, mark size=1.6pt, solid, mark repeat=1,mark options=solid]
file{data/set_b_3x128_no_filler_sum.txt};
\addplot[color=red, mark=o, mark size=1.6pt, solid, mark repeat=1,mark options=solid]
file{data/set_b_3x768_no_filler_sum.txt};
\end{axis}%
\end{tikzpicture}%
}%
    }%
    \caption{Precision-recall plot for (a) Set A (multisyllabic keywords) and (b) Set B (monosyllabic keywords) with various RNN sizes.}
    \label{fig:plot_sizes}
\end{figure*}

We now compare the proposed keyword spotters with the networks sizes of 3x128 and 3x768 as in \figurename~\ref{fig:plot_sizes}. The decoder model is the keyword-only model without employing the sum-to-max approximation. The maximum F1 scores for the baseline DNN-HMM, small CTC (3x128), and large CTC (3x768) models are 0.936, 0.964, and 0.980 for Set A, and 0.517, 0.806, and 0.847, respectively, for Set B. Note that the number of parameters in the front-end networks for the baseline, small CTC, and large CTC models are 6.61 M, 0.40 M, and 12.21 M, respectively. Considering that the number of parameters is approximately proportional to the number of arithmetic operations per frame in the front-end networks, the proposed keyword spotter with the network size of 3x128 outperforms the baseline DNN-HMM model with only 6\% of computations. Moreover, the back-end decoder network is much simpler in the proposed keyword-only model since the number of state transitions is significantly smaller than those of the triphone-based HMM keyword-filler model.

\section{Concluding Remarks}
\label{sec:conclusion}

Throughout the paper, an end-to-end RNN-based context-aware dictionary-free keyword spotter is proposed. Due to the weak language model embedded in the front-end RNN, the proposed keyword spotter shows remarkable performance improvement over the baseline DNN-HMM keyword spotter even with significantly less computations. Furthermore, the back-end of the proposed one is much simpler than that of the DNN-HMM model. Therefore, the proposed CTC keyword spotter is suitable for small-footprint low-latency applications with continuous speech input.


%



\section*{Acknowledgment}

This work was supported in part by the Brain Korea 21 Plus Project and the National Research Foundation of Korea (NRF) grants funded by the Korea government (MSIP) (No. 2015R1A2A1A10056051).


\ifCLASSOPTIONcaptionsoff
  \newpage
\fi

\pagebreak



\bibliographystyle{IEEEtran}
\bibliography{IEEEabrv,refs}

%








\end{document}